\documentclass[11pt,letterpaper,logo,onecolumn]{microsoft}
\usepackage{natbib}
\usepackage{amsfonts}
\usepackage{subcaption}
\usepackage{pgfplots}
\pgfplotsset{compat=1.18}

\pgfplotsset{
  msgroup/.style={
    ybar,
    width=0.88\linewidth,
    height=5.4cm,
    bar width=8pt,
    enlarge x limits=0.14,
    nodes near coords,
    ylabel style={font=\small},
    tick label style={font=\footnotesize},
    legend style={font=\footnotesize, draw=mslightgray, fill=white,
        at={(0.5,1.02)}, anchor=south, legend columns=-1,
        /tikz/every even column/.append style={column sep=7pt}},
    axis lines*=left,
    ymajorgrids, grid style={draw=mslightgray},
    xtick style={draw=none},
    clip=false,
  },
  msrank/.style={
    xbar,
    width=0.70\linewidth,
    bar width=6.5pt,
    y=0.60cm,
    enlarge y limits={abs=0.8},
    nodes near coords,
    nodes near coords style={font=\scriptsize,
        /pgf/number format/fixed, /pgf/number format/precision=1},
    every node near coord/.append style={anchor=west, xshift=1.5pt},
    xlabel style={font=\small},
    tick label style={font=\footnotesize},
    axis lines*=left,
    xmajorgrids, grid style={draw=mslightgray},
    ytick style={draw=none},
    clip=false,
  },
}

\title{Navigation-Informed Embeddings: Dense-Retriever Adaptation from Agent Search Traces}

\author{Shrey Shah and Levent Ozgur}

\date{}

\begin{document}

\begin{abstract}
Agentic retrieval workflows produce query, retrieval, and stopping traces as a byproduct of answering questions. We study how these traces can adapt a deployed dense retriever to changing workflow distributions without new relevance labels, synthetic queries, or LLM judgments. We introduce \emph{Navigation-Informed Embeddings} (NIE), a family of trace-derived objectives. NIE-Stop turns the stopping document into a soft positive; NIE-Path additionally uses preceding path documents as hard comparisons and imposes ordinal constraints with geometric decay. A BGE encoder adapted from retained source trajectories improves support Recall@20 on an independent target benchmark from 72.2 to 78.0 overall. NIE-Stop reaches 76.9 overall and 52.3 on long paths; NIE-Path raises long-path performance to 55.4, compared with 46.7 for the unadapted encoder. A shuffled-order control under the full path objective loses 3.2 points. Without public-benchmark training, the same adapter also improves nDCG@10 by 1.9 points on standard BEIR HotpotQA. NIE therefore provides a lightweight adaptation channel for settings where trajectories are already retained, with zero incremental labeling cost.
\end{abstract}

\maketitle

\section{Introduction}

Document retrieval systems must identify relevant documents from large collections given user queries. While dense retrieval methods have achieved strong results on standard benchmarks \cite{karpukhin2020dense, khattab2020colbert}, adapting them to specialized corpora typically requires relevance judgments or curated query-document pairs. Deployed agentic retrieval workflows face an additional problem: query and corpus distributions evolve after the retriever is trained, while producing a new labeled adaptation set for each shift is expensive.

LLM-based question-answering systems already generate a source of adaptation supervision: \emph{agent navigation traces}. When an agent navigates through multiple documents to answer a question, the sequence records more than a set of viewed documents. It records a stopping event and a refinement process: broad or partially relevant documents are often visited early, while later documents tend to address the information need more directly. Agents can overshoot, stop on incomplete evidence, or continue searching after seeing useful context, so the signal is best modeled as soft behavioral evidence. The central question of this paper is whether that evidence can update the dense retrieval component as workflow distributions change.

Consider the spectrum of LLM search behaviors. Some questions are resolved immediately---the agent retrieves a single document and finds the answer. Others require moderate exploration, with the agent viewing two or three documents. Still others involve extended search paths where the agent navigates through many documents before reaching the answer. Treating each document as an independent relevance judgment discards this structural information entirely.

We propose \emph{Navigation-Informed Embeddings} (NIE), a framework that adapts document representations from LLM agent search paths. NIE applies when search trajectories are already retained as part of an agentic retrieval workflow. Its training inputs are the observed query, the ordered retrieved documents, and the stopping point; human labels are reserved for evaluation. We study two nested forms of supervision:

\begin{enumerate}
\item \textbf{Terminal documents provide positive signal.} When an agent stops searching after retrieving a document, that document often contains information sufficient to answer the query, even though individual terminal documents may be noisy.

\item \textbf{Path order encodes relevance gradients.} Documents retrieved later in a search session tend to be more relevant than those retrieved earlier, as agents progressively refine their search toward the answer. This ordering provides richer supervision than treating the path as an unordered bag.
\end{enumerate}

NIE converts agentic retrieval traces into a dense-retriever adaptation set. \emph{NIE-Stop} uses only the stopping document from each trajectory. \emph{NIE-Path} extends it with nonterminal path documents and chronology for the extended searches where path structure is available.

This work makes the following contributions:
\begin{itemize}
\item We formulate post-deployment dense-retriever adaptation from agent trajectories. NIE-Stop converts stopping events into contrastive supervision without new labeling or generation calls.
\item We introduce NIE-Path, which extends terminal-only adaptation with path-derived hard comparisons, an adjacent ordering loss, and geometric decay.
\item We evaluate under a held-out target shift: training uses retained source trajectories, while retrieval quality is measured on an independent target benchmark with human supporting-evidence judgments. NIE improves overall support Recall@20 by 5.8 points and long-path recall by 8.7 points over its dense initializer.
\item We isolate the available supervision signals with NIE-Stop, shuffled-order, reversed-order, random-terminal, and negative-source controls, and test external transfer on HotpotQA.
\end{itemize}
Together, these results establish the supervision source for an adaptation loop in which each encoder update uses newly retained agent search traces with zero incremental labeling cost.

\section{Related Work}
\label{sec:related}
\paragraph{Learning from Implicit Feedback.}
Click-through data has long been used for learning to rank \cite{joachims2002optimizing, agichtein2006improving}.
Session-based approaches model query reformulation and document access patterns \cite{ahmad2019context}.
Listwise ranking methods \cite{cao2007learning} optimize over document orderings.
A key challenge in learning from implicit feedback is systematic bias, especially presentation and position bias, motivating counterfactual and propensity-weighted approaches for unbiased learning-to-rank \cite{joachims2017unbiased}. NIE inherits the broad implicit-feedback premise but changes the unit of supervision. Classical click-based ranking typically aggregates heterogeneous users' reactions to displayed slates and must infer relevance from clicks, skips, dwell time, and examination propensities. NIE instead observes an active information-seeking episode conditioned on a fixed query and goal: the agent selects documents through iterative retrieval calls, consumes the returned evidence, and then continues searching or answers. This trajectory-conditioned continuation signal provides behavioral evidence about the usefulness of documents along the path.

\paragraph{Agent Trajectory Feedback for Dense Adaptation.}
NIE instantiates implicit feedback at the level of agent trajectories rather than user interactions with a static result list. In a displayed-result setting, a skipped item may be unseen; in an agent trajectory, an earlier item was explicitly retrieved into the agent's context but was insufficient for stopping. The signal is also component-level: it adapts a dense encoder that can be reused wherever that encoder serves as a retrieval scorer. This setting induces a specific data-construction problem, mapping each trajectory into terminal positives, path-local contrasts, and adjacent order constraints for dense representation learning.
\paragraph{LLM-Generated Synthetic Supervision for Retrieval.}
Recent work uses large language models to synthesize retrieval supervision by generating queries for documents and training retrievers on the resulting synthetic pairs or triples \cite{bonifacio2022inpars, dai2022promptagator}.
NIE instead extracts supervision from observed agent navigation behavior and leverages the induced ordering along search paths as a soft relevance gradient for training.
\paragraph{Dense Passage Retrieval.}
DPR \cite{karpukhin2020dense} introduced dual-encoder architectures for open-domain question answering, training on question-passage pairs with in-batch negatives.
Subsequent work improved training through hard negative mining \cite{xiong2020approximate}, knowledge distillation \cite{hofstatter2020improving}, and contrastive pre-training \cite{gao2021simcse, izacard2021contriever}.
ColBERT \cite{khattab2020colbert} introduced late interaction for improved effectiveness.
These methods are largely agnostic to how supervision is obtained, and can be trained using human annotations, synthetic supervision, or implicit feedback.
Our work extends this line by replacing explicit supervision with implicit signals from LLM agent navigation.
\paragraph{Multi-Hop Retrieval.}
HotpotQA \cite{yang2018hotpotqa}, 2WikiMultiHopQA \cite{ho2020constructing}, and MuSiQue \cite{trivedi2022musique} introduced benchmarks requiring reasoning across multiple documents.
MDR \cite{xiong2021answering} performs iterative retrieval conditioned on previously retrieved passages, while Baleen \cite{khattab2021baleen} uses condensed retrieval for multi-hop reasoning.
PathRetriever \cite{asai2020learning} learns to follow reasoning chains over knowledge graphs.
Our work differs in exploiting navigation sequences for \emph{training} rather than inference-time reasoning.

\section{Method}
\label{sec:method}

\subsection{Problem Setting}

Let $\mathcal{C} = \{d_1, \ldots, d_N\}$ denote a corpus of documents. Given a query $q$, the retrieval task is to identify the top-$k$ most relevant documents from $\mathcal{C}$. We assume access to LLM agent navigation traces: for each query, we observe the sequence of retrieval steps and documents the agent retrieved before producing an answer. NIE builds the training objective from these behavior traces, while human judgments are used for held-out evaluation.

\subsection{Implicit Supervision from Agent Search Traces}

We extract training signal directly from agent navigation traces without any explicit labeling. For each search session with query $q$ and navigation path $P = (d_1, d_2, \ldots, d_T)$, NIE constructs three kinds of weak supervision:

\paragraph{Soft positives.} The terminal document $d_T$---the final document retrieved before the agent stopped making retrieval calls and produced an answer---serves as a \emph{soft positive} for query $q$. This soft label captures the empirical tendency for terminal documents to be more useful than earlier documents in the path. NIE does not require access to a calibrated stopping score; the observed termination event is the supervision signal.

\paragraph{Lower-ranked path items.} Earlier documents in the path $(d_1, \ldots, d_{T-1})$ serve as lower-ranked path items. These are documents the agent retrieved but continued past. They may contain partial or even necessary supporting information, while the path provides weak evidence that they were less useful for the final answer than the terminal document. NIE uses these documents as trajectory-derived comparison points: they provide challenging alternatives to the terminal document and define the local order relationships used by the path-wise objective introduced below.

\paragraph{Random negatives.} Additional negatives are sampled from the corpus, providing contrast against documents that are unlikely to be related to the query.

This formulation differs from binary supervision in an important respect: the navigation path provides a \emph{preference ordering} rather than absolute labels. We model $d_T \succ d_{T-1} \succ \cdots \succ d_1$, where $\succ$ denotes ``more useful to the agent's final answer than.'' Earlier documents can be genuinely relevant, especially for multi-hop questions; NIE treats them as lower-confidence positives or lower-ranked contrast documents depending on the objective.

\paragraph{Policy dependence.} The induced preferences are conditional on the behavior of the agent that generated the trajectory. If that agent repeatedly misses relevant evidence, overuses a narrow retrieval pattern, or stops prematurely, trajectory-derived supervision can reflect those tendencies. This motivates held-out human evaluation and the behavioral-noise analysis below.

\paragraph{Trajectory-only training objective.} NIE constructs training pairs from the observed query, the ordered path, the stopping document, and unlabeled corpus documents used as random negatives. LLM-as-judge labels, human relevance labels, and answer-derived labels are reserved for evaluation and auditing.

\paragraph{Worked schematic example.}
Consider the query ``Which regulator approved the product, and what monitoring was required?'' Table~\ref{tab:worked_trajectory} shows a schematic four-step trajectory and the supervision NIE constructs from it.

\begin{table}[t]
\centering
\small
\caption{Schematic conversion of an agent trajectory into NIE supervision.}
\label{tab:worked_trajectory}
\begin{tabular}{@{}p{0.07\textwidth}p{0.47\textwidth}p{0.38\textwidth}@{}}
\toprule
Step & Retrieved document and agent action & NIE supervision \\
\midrule
$d_1$ & Broad product overview; continue searching & Lower-ranked path contrast \\
$d_2$ & Manufacturer announcement; continue searching & Contrast; $d_2 \succ d_1$ \\
$d_3$ & Approval notice naming the regulator; continue searching & Contrast; $d_3 \succ d_2$ \\
$d_4$ & Safety notice specifying the monitoring rule; answer and stop & Soft positive; $d_4 \succ d_3$ \\
$d^-$ & Unrelated document sampled from the corpus & Random negative \\
\bottomrule
\end{tabular}
\end{table}

The ordering prior is well matched to progressive refinement, where early documents are broad and later documents resolve increasingly specific parts of the query. It is imperfect under evidence accumulation: here $d_3$ may remain necessary for identifying the regulator even though the agent continues to $d_4$. NIE therefore interprets $d_4 \succ d_3$ as a behavioral preference within the observed episode, not as a claim that $d_3$ is irrelevant. Adjacent comparisons, geometric decay, and support-recall evaluation limit and measure the consequences of this approximation.

\subsection{Terminal-Only Contrastive Training}

We use the terminal document as the positive in a standard contrastive objective:
\begin{equation}
\mathcal{L}_{\text{contrast}} = -\log \frac{\exp(\text{sim}(q, d^+) / \tau)}{\sum_{d \in \mathcal{B}} \exp(\text{sim}(q, d) / \tau)}
\end{equation}
where $\tau$ is a temperature parameter and $\mathcal{B}$ contains one terminal positive together with random corpus negatives and positives from other examples in the batch. We call this objective \emph{NIE-Stop}: it excludes every nonterminal path document and uses neither path order nor geometric decay. NIE-Path augments the contrastive set with earlier path documents as hard comparisons and, for long paths, adds the path-ordering loss below. We use ``contrast'' to denote a lower-ranked item relative to the stopping point under the observed trajectory.

\subsection{Path-Wise Training for Extended Search Paths}

Not all documents along a navigation path provide equally informative supervision. Documents accessed earlier often reflect exploratory or coarse retrieval, while later documents reflect increasingly refined relevance. We therefore assign position-dependent relevance weights that decay with distance from the terminal document, capturing diminishing confidence in earlier positions. For queries where agents navigated through four or more documents, we model a navigation path $P = (d_1, d_2, \ldots, d_T)$ as an ordered sequence with increasing expected usefulness toward the terminal document $d_T$.

\paragraph{Geometric Relevance Decay.} Relevance weights follow a geometric decay:
\begin{equation}
w_t = \gamma^{T-t}
\end{equation}
where $t$ indexes position, $T$ is the path length, and $\gamma \in (0, 1)$ controls the decay rate. This assigns weight 1 to the terminal document and geometrically decreasing weights to predecessors.

\paragraph{Ordering Loss.} The ordering loss encourages embeddings to respect the relevance ordering:
\begin{equation}
\mathcal{L}_{\text{order}} = \frac{1}{T-1}\sum_{t=1}^{T-1} \max\bigl(0, \text{sim}(q, d_t) - \text{sim}(q, d_{t+1}) + m \cdot w_{t+1}\bigr)
\end{equation}
where $m$ is a margin hyperparameter. Violations, where an earlier document embeds closer to the query than a later one, incur a hinge penalty; the geometric factor scales the required margin rather than multiplying the loss after the hinge. Normalizing by $T-1$ prevents long paths from dominating the batch solely because they contain more adjacent pairs. The loss compares adjacent steps rather than all document pairs, which reduces the penalty when early documents provide legitimate supporting context.

\paragraph{Combined Objective.} The full path-wise objective combines ordering and contrastive losses:
\begin{equation}
\mathcal{L}_{\text{path}} = \mathcal{L}_{\text{contrast}} + \lambda \mathcal{L}_{\text{order}}
\end{equation}
where the contrastive term ensures the terminal document ranks above negatives, and the ordering term encourages the embedding geometry to preserve path structure. We use $\lambda=1$ in all experiments. In the full model, short and moderate paths contribute the augmented contrastive term; long paths contribute both terms.

\section{Experimental Setup}
\label{sec:setup}

\paragraph{Target workflow benchmark.} Our primary experiments use an interaction corpus from agentic retrieval workflows. This benchmark captures the adaptation regime NIE targets: source trajectories provide supervision, the target query distribution has shifted, and retrieval quality is measured against independently annotated supporting evidence. The target benchmark is challenging for off-the-shelf dense retrieval: BGE-base obtains 46.7 support Recall@20 on long-path target queries, compared with 61.3 Recall@20 for recovering both HotpotQA supports under the setup in Section~\ref{sec:hotpotqa}.

\paragraph{Workflow and measured shift.} Source and target were collected in successive three-month windows after a task-requirements update. Both windows use the same fixed agent model, prompt template, retrieval tools, and hybrid candidate-generation stack; neither uses an NIE-adapted checkpoint. Sessions ending at the tool-call limit or through execution failure are excluded. Exact and MinHash-based near-duplicate checks find no query or document overlap across corpora. Relative to source validation, the target window has a higher fraction of four-or-more-step trajectories (36\% versus 27\%) and lower unadapted BGE support Recall@20 (72.2 versus 81.0), providing operational and retrieval-level measures of the shift.

\paragraph{Source navigation traces.} We use query chains from LLM agent search sessions over the source corpus. Each training example contains the query, the ordered retrieved documents, and the document at which the agent stopped. We categorize trajectories by search path length: \emph{short paths} where the answer was found after viewing one document, \emph{moderate paths} requiring two to three documents viewed, and \emph{long paths} with four or more documents viewed. The source set includes approximately 1,500 queries with extended search paths for path-wise training. Qualitative inspection indicates that many long-path queries involve refinement across multiple partially useful documents rather than a single obvious lookup.

\paragraph{Retained trajectory records.} Source and target trajectories are retained records of completed agentic retrieval sessions. The workflow follows a tool-use pattern: given a query and current context, the agent may issue retrieval calls to available search tools, observe returned documents, and either issue another retrieval call or generate a final answer. The stopping document is the final retrieved document before answer generation. Thus, stopping is the model choosing to answer rather than invoke another retrieval tool, not an externally calibrated relevance judgment or a separate stopping classifier. The trajectories were generated by the existing agent workflow before NIE training and did not use the adapted BGE checkpoint. Each record exposes the fields used by NIE: the original user query, retrieval-step buckets, document order within the session, and the stopping document. Retrieval calls issued in the same agent step remain in a common bucket, and NIE imposes no order among documents in that bucket. NIE embeds the original user query; step-specific tool queries and accumulated agent context are not training inputs. At evaluation time, each dense encoder retrieves over the target corpus independently; trajectory path length is used only for stratification.

For an audit of implicit-signal quality, four annotators independently assessed a random sample of 1,000 terminal documents from the 2,197 trajectory-derived training examples. Each annotator made a binary judgment of whether the terminal document contained the correct evidence for the associated query. We marked a terminal document correct when at least three of four annotators judged it correct; ties and lower vote counts were marked wrong. Under this aggregation rule, 26\% of sampled terminal documents were wrong. These audit judgments are used only to characterize the supervision signal: they do not filter examples or enter training, model selection, or target evaluation.

\paragraph{Dataset size.} The target retrieval corpus contains 47,109 documents. Documents contain 205 words on average (median 185; range 7--647). The source split contains 2,484 training queries and 529 validation queries, producing 2,197 query-level training examples after trace extraction. The held-out target evaluation contains 542 queries, with 191, 156, and 195 queries in the three reported evaluation buckets, respectively.

\paragraph{Held-out target evaluation.} The main evaluation uses an independent target benchmark with held-out queries, curated ground-truth answers, associated supporting evidence, and human relevance judgments. These supporting-evidence records, rather than target stopping behavior, define retrieval relevance. NIE training and model selection use no target queries, documents, or relevance labels. Source and target have disjoint queries and document identifiers and were collected independently under evolving task requirements, while being produced by the same class of agentic retrieval workflow. The experiment therefore measures one source-to-target encoder update in the workflow family. Applying the same construction to successive windows of retained trajectories provides a direct mechanism for continuous adaptation without a new labeling interface.

The benchmark's curated answer/evidence records supply the target support sets directly. We use these records as ground-truth relevance labels and determine retrieval correctness by matching retrieved documents against them. We do not derive, filter, or relabel target supports from the target trajectories, and target labels are never used for training or source validation.

\paragraph{Train/test separation.} The training objective uses source trajectory-derived signals: terminal documents, earlier path documents, path order, and random source-corpus negatives. For training, the document at which the agent stopped is treated as a soft positive, and earlier documents provide path-derived contrast and order signal. Human relevance annotations and curated answer/evidence records are used for evaluation. Training trajectories share no queries or documents with validation or target evaluation records.

\paragraph{Path-length stratification.} Path length is assigned from the trajectory-generating agent's original search path before training or evaluation. For target-domain evaluation, these trajectories define analysis buckets; their terminal documents and path order are excluded from training supervision. All retrievers are evaluated on the same query buckets. Thus, path length is an operational difficulty measure for the agent that produced the trajectory. We use the buckets for analysis and reporting rather than as additional input features to the retriever.

\paragraph{Evaluation labels.} On the target-domain test set, we report support Recall@20, stratified by observed path length. For query $i$ with annotated support set $G_i$ and retrieved top-20 set $R_i^{20}$, the per-query score is $|G_i \cap R_i^{20}|/|G_i|$; reported values average this fraction across queries. Evaluation relevance is defined by human-annotated supporting-evidence records rather than by the target agent's stopping behavior. Target trajectories are used only to stratify queries by observed path length; their terminal documents and path order are excluded from training supervision. The curated answer/evidence dataset measures retrieval quality, while NIE training is built from retained source trajectories. For HotpotQA \cite{yang2018hotpotqa}, we separately report the fraction of questions for which both annotated supporting passages or either passage appears in the top 20.

\paragraph{Dense adaptation protocol.} NIE adapts an existing dense encoder. The empirical question is therefore whether trajectory-derived supervision improves that encoder under target distribution shift. We center the evaluation on the BGE-base initializer and its trajectory-adapted variants, with additional off-the-shelf dense encoders as reference points:
\begin{itemize}
\item BGE-base-en-v1.5 \cite{bge_embedding}: the adapted dense initializer before trajectory training.
\item Contriever \cite{izacard2021contriever} and E5-base \cite{wang2022text}: off-the-shelf dense encoders included as additional reference models.
\item NIE-Stop: the same BGE initializer fine-tuned on terminal-document positives with random and in-batch negatives, excluding all nonterminal path documents.
\item NIE-Path: NIE-Stop augmented with nonterminal path documents as hard comparisons, adjacent path-order constraints, and geometric weighting; shuffled, reversed, and random-terminal variants perturb one trajectory signal under the same training recipe.
\end{itemize}

The comparison is component-level: all NIE variants retrieve with the same dense nearest-neighbor interface, so differences measure trajectory-derived adaptation of the component updated by NIE. Sparse, late-interaction, and hybrid systems address complementary system-level design choices; the adapted encoder can be used directly or as the dense scorer inside those stacks.

\paragraph{External transfer.} We also evaluate the already-trained adapter on the HotpotQA development set from the distractor release. Each question-context paragraph is represented by its title and text, and the context paragraphs are pooled into a global retrieval corpus rather than ranked only within each question's ten candidates. Repeated paragraphs associated with different questions retain question-specific identifiers. Gold passages are matched by these identifiers, so retrieving identical content attached to another question does not receive credit and duplicate instances can compete within the top 20. Because identical model inputs can carry different identifiers, absolute scores depend partly on duplicate competition and tie handling; all models are compared on the same fixed corpus and labels. We refer to this construction as \emph{pooled HotpotQA} to distinguish it from both the standard per-question distractor task and full-Wikipedia retrieval. HotpotQA questions and supporting-fact labels do not enter model fitting. The experiment is a secondary transfer diagnostic with independently annotated multi-hop supports.

\paragraph{Role of each evaluation.} The target workflow benchmark evaluates the intended distribution-shift setting: adapting an existing dense scorer from pre-existing traces without target-domain labels. HotpotQA evaluates cross-benchmark transfer on independently constructed multi-hop data. Together, these evaluations distinguish in-workflow adaptation from transfer beyond the primary evaluation pipeline.

\paragraph{Implementation.} We fine-tune all parameters of BGE-base-en-v1.5 using AdamW with weight decay 0.01. Hyperparameters are selected on source validation Recall@20 computed from held-out trajectory terminal documents. Training runs for three epochs and uses in-batch negatives, earlier path documents as lower-ranked comparisons, and randomly sampled corpus negatives. No LLM or human relevance judgments are used to choose positives or negatives for training. Similarity is computed with the same normalized embedding score used by the BGE initializer. The full model weights the contrastive and ordering terms equally in the combined objective. Path-wise training uses $\gamma=0.7$ and margin $m=0.1$.

\section{Results}
\label{sec:results}

\subsection{Main Results}

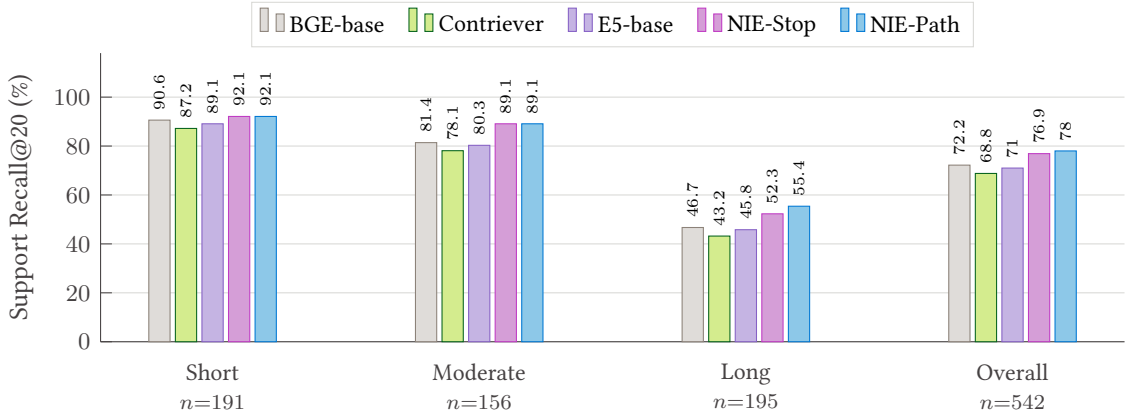
\begin{figure}[t]
\centering
\begin{tikzpicture}
\begin{axis}[
  msgroup,
  ymin=0, ymax=118,
  ytick={0,20,40,60,80,100},
  ylabel={Support Recall@20 (\%)},
  symbolic x coords={Short, Moderate, Long, Overall},
  xtick=data,
  xticklabels={{Short\\{\scriptsize $n{=}191$}},{Moderate\\{\scriptsize $n{=}156$}},%
               {Long\\{\scriptsize $n{=}195$}},{Overall\\{\scriptsize $n{=}542$}}},
  xticklabel style={align=center},
  nodes near coords style={font=\tiny, rotate=90, anchor=west,
      /pgf/number format/fixed, /pgf/number format/precision=1},
]
\addplot[draw=msdata8, fill=msfill8] coordinates {(Short,90.6) (Moderate,81.4) (Long,46.7) (Overall,72.2)};
\addplot[draw=msdata6, fill=msfill6] coordinates {(Short,87.2) (Moderate,78.1) (Long,43.2) (Overall,68.8)};
\addplot[draw=msdata4, fill=msfill4] coordinates {(Short,89.1) (Moderate,80.3) (Long,45.8) (Overall,71.0)};
\addplot[draw=msdata2, fill=msfill2] coordinates {(Short,92.1) (Moderate,89.1) (Long,52.3) (Overall,76.9)};
\addplot[draw=msdata1, fill=msfill1] coordinates {(Short,92.1) (Moderate,89.1) (Long,55.4) (Overall,78.0)};
\legend{BGE-base, Contriever, E5-base, NIE-Stop, NIE-Path}
\end{axis}
\end{tikzpicture}
\caption{Target-domain support Recall@20 by observed trajectory path length. NIE-Stop uses only stopping documents from the trajectories; NIE-Path uses the complete path objective. The two NIE variants coincide on short and moderate queries after rounding; they separate on long paths.}
\label{fig:main_results}
\end{figure}

Figure~\ref{fig:main_results} presents target-domain support Recall@20 across query types.

NIE improves support Recall@20 by 1.5 points on short-path queries, 7.7 points on moderate-path queries, and 8.7 points on long-path queries relative to the BGE-base baseline. Across all 542 target queries, the query-count-weighted score increases from 72.2 to 78.0, a 5.8-point gain. The strongest gains occur on queries that required longer agent searches, suggesting that trajectories expose useful-evidence signals not captured by the original embedding model.

The two NIE variants expose how much of the retained trace is used. NIE-Stop reaches 76.9 overall and improves long-path recall by 5.6 points using only stopping documents. NIE-Path raises overall recall to 78.0 and long-path recall by a further 3.1 points by adding nonterminal path comparisons, adjacent ordering, and geometric decay. The 3.1-point increment therefore measures the complete path objective beyond terminal-only adaptation rather than the ordering term in isolation. Both variants are trained as single global retrievers, and their short and moderate results are equal after rounding.

Since source trajectories and the main evaluation benchmark are separate, these gains reflect trajectory-derived adaptation under a no-target-label setting rather than in-domain fine-tuning on the evaluation corpus.

The moderate and long-path gains over the initializer are the primary empirical effect. Because support Recall@20 averages the fraction of annotated supports recovered for each query, it measures evidence coverage rather than crediting a query solely for retrieving one relevant document. Performance improves most in the moderate and long-path buckets, where the risk of suppressing multi-document evidence is most salient; systematic demotion of required supports would lower this metric.

\subsection{Matched Trace-Adaptation Baselines}

These controls test whether the gains arise from generic adaptation on the retained records or from the stopping and path structure used by NIE. Every method starts from the same BGE initializer and uses the same source-query budget, optimizer schedule, and source validation split; none uses human or LLM judgments. ANN self-training pairs each original source query with the initializer's top retrieved document. All-viewed-positive training pairs the original query with every document in its trajectory as unordered positives, removing both the stopping distinction and chronology. Tool-query training instead uses each intermediate retrieval query issued by the agent paired with the document returned for that call. NIE-Stop pairs the original query only with its stopping document. The matched no-order control uses the same stopping positives, nonterminal comparisons, random negatives, and in-batch negatives as NIE-Path but sets the ordering-loss coefficient to zero. NIE-Path then adds the observed adjacent chronology and geometric weighting.

\begin{figure}[t]
\centering
\begin{tikzpicture}
\begin{axis}[
  msrank,
  xmin=44, xmax=84,
  xtick={45,50,...,80},
  ytick={1,...,7},
  yticklabels={
    {\textbf{NIE-Path}},
    {NIE-Path, no ordering loss},
    {NIE-Stop},
    {Retained tool-query pairs},
    {All viewed documents positive},
    {ANN self-training},
    {BGE-base (no adaptation)}},
  xlabel={Support Recall@20 (\%)},
  legend style={font=\footnotesize, draw=mslightgray, fill=white,
      at={(0.5,-0.30)}, anchor=north, legend columns=-1,
      /tikz/every even column/.append style={column sep=7pt}},
]
\addplot[xbar, bar shift=-4pt, draw=msdata1, fill=msfill1] coordinates {
  (46.7,7) (50.2,6) (51.5,5) (51.9,4) (52.3,3) (52.8,2) (55.4,1)};
\addplot[xbar, bar shift=4pt, draw=msdata8, fill=msfill8] coordinates {
  (72.2,7) (74.8,6) (75.5,5) (75.9,4) (76.9,3) (77.1,2) (78.0,1)};
\legend{Long-path, Overall}
\end{axis}
\end{tikzpicture}
\caption{Matched trace-adaptation baselines on the target benchmark, ordered by
long-path support Recall@20. Every method starts from the same BGE initializer
under an identical training budget.}
\label{fig:matched_adaptation}
\end{figure}
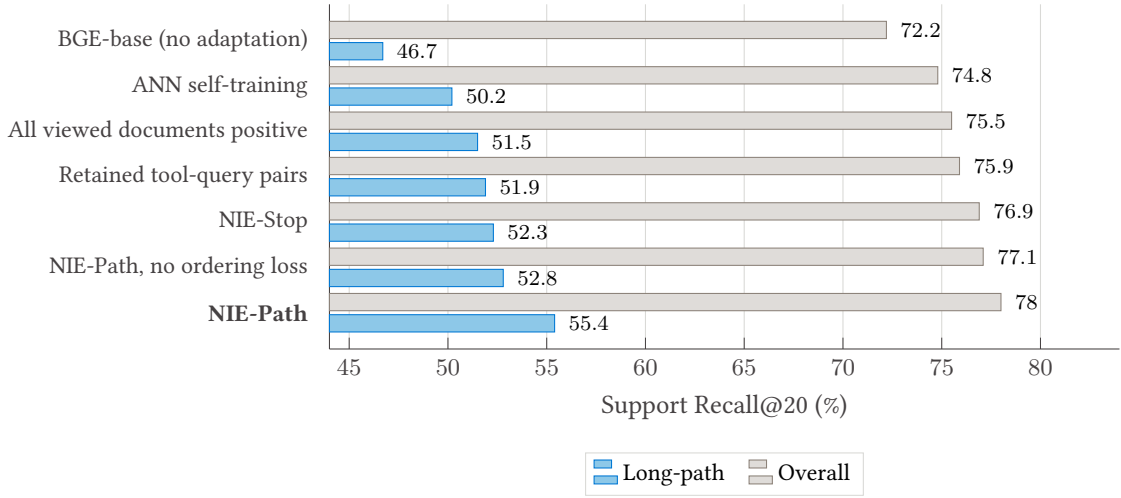

All trace-derived alternatives improve over the unadapted initializer, showing that several label-free uses of the retained records provide adaptation signal. The strongest generic construction, retained tool-query pairs, reaches 51.9 long-path and 75.9 overall Recall@20; using the stopping event in NIE-Stop reaches 52.3 and 76.9. Adding nonterminal comparisons without ordering reaches 52.8 and 77.1. NIE-Path then improves over this matched no-order control by 2.6 points on long paths and 0.9 points overall. Figure~\ref{fig:matched_adaptation} therefore separates generic trace adaptation, stopping-document supervision, path-derived comparisons, and observed chronology under a common initializer and training budget.

\subsection{Ablation Study}

\begin{figure}[t]
\centering
\begin{tikzpicture}
\begin{axis}[
  width=0.72\linewidth,
  xbar,
  xmin=45.5, xmax=57.4,
  bar width=7.5pt,
  y=0.575cm,
  enlarge y limits={abs=0.75},
  ytick={1,...,10},
  yticklabels={
    {w/o in-batch negatives},
    {w/o path-derived hard neg.},
    {Random terminal document},
    {Reversed path order},
    {Shuffled path order},
    {\textbf{NIE-Path (full)}},
    {w/o geometric decay},
    {No ordering loss ($\lambda{=}0$)},
    {NIE-Stop},
    {BGE-base (no adaptation)}},
  nodes near coords,
  nodes near coords style={font=\footnotesize,
      /pgf/number format/fixed, /pgf/number format/precision=1},
  every node near coord/.append style={anchor=west, xshift=1.5pt},
  xlabel={Long-path support Recall@20},
  xlabel style={font=\small},
  tick label style={font=\footnotesize},
  axis lines*=left,
  xmajorgrids, grid style={draw=mslightgray},
  ytick style={draw=none},
  clip=false,
]
\addplot[xbar, bar shift=0pt, draw=msdata1, fill=msfill1] coordinates {
  (46.7,10) (52.3,9) (52.8,8) (54.1,7) (55.4,6)};
\addplot[xbar, bar shift=0pt, draw=msdata3, fill=msfill3] coordinates {
  (52.2,5) (48.7,4) (47.1,3)};
\addplot[xbar, bar shift=0pt, draw=msdata8, fill=msfill8] coordinates {
  (51.8,2) (49.2,1)};
\end{axis}
\end{tikzpicture}
\caption{Ablation study and path-order controls on long-path queries (4+ documents
viewed). Objective components (blue) build from the unadapted initializer to full
NIE-Path; trajectory-order controls (orange) perturb one trajectory signal under the
same recipe; negative-source ablations (gray) vary the contrast set. NIE-Path adds
3.1 support Recall@20 beyond terminal-only adaptation.}
\label{fig:ablation}
\end{figure}
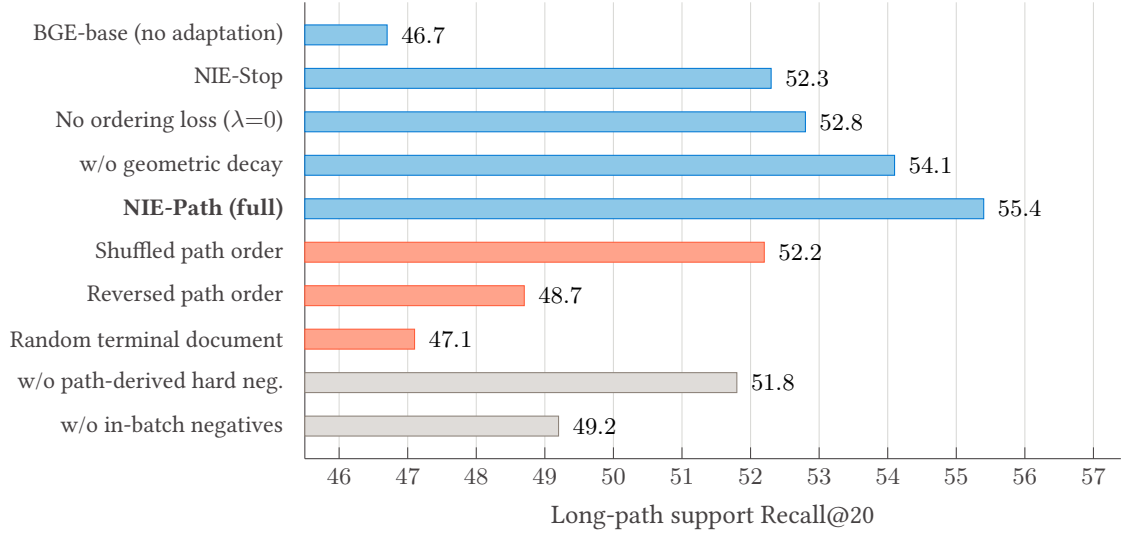

Figure~\ref{fig:ablation} analyzes the contribution of each component and tests whether the observed path order is carrying useful signal.

NIE-Stop improves long-path support Recall@20 by 5.6 points using terminal positives without nonterminal path documents. The matched no-order control uses the same terminal positives, nonterminal path documents as hard comparisons, and random and in-batch negatives as NIE-Path, but sets $\lambda=0$; it reaches 52.8. Adding adjacent ordering with an unweighted margin reaches 54.1, and geometric weighting increases performance to 55.4. Full NIE therefore adds 2.6 points over the matched no-order objective. The shuffled, reversed, and random-terminal controls use the NIE-Path training recipe. In the shuffled and reversed controls, the observed stopping document remains fixed as the terminal positive; only the order of nonterminal path documents is permuted or reversed. Thus, NIE-Path and its shuffled-order control use the same terminal documents, path documents, contrast sets, and decay, differing only in the chronology assigned to nonterminal path items. The resulting 3.2-point gap isolates observed chronology relative to arbitrary nonterminal order under the matched objective. Reversing nonterminal order (48.7) and randomizing the terminal document in the separate random-terminal control (47.1) further reduce performance. The negative-source rows isolate contrast-set composition within the full objective: excluding path-derived hard negatives yields 51.8, while excluding in-batch negatives yields 49.2.

We estimate uncertainty with 10,000 paired bootstrap samples over target queries. On the long-path bucket, the BGE-to-NIE-Stop gain is $+5.6$ points (95\% CI $[2.1, 9.0]$), the matched no-order-to-NIE-Path gain is $+2.6$ points ($[0.5, 4.8]$), and the observed-order advantage over shuffled order is $+3.2$ points ($[0.9, 5.6]$). Across three training seeds, long-path support Recall@20 is $52.8 \pm 0.5$ for the matched no-order control and $55.4 \pm 0.4$ for NIE-Path (mean $\pm$ standard deviation).

\subsection{Effect of Path Length}

Performance varies with the length of navigation paths in the test set:
\begin{figure}[t]
\centering
\begin{tikzpicture}
\begin{axis}[
  msgroup,
  width=0.66\linewidth, height=5.0cm,
  ymin=0, ymax=68,
  ylabel={Long-path support Recall@20 (\%)},
  symbolic x coords={A,B,C},
  xtick=data,
  xticklabels={{4--6 docs},{7--12 docs},{13+ docs}},
  bar width=15pt,
  enlarge x limits=0.28,
  nodes near coords style={font=\scriptsize, anchor=south,
      /pgf/number format/fixed, /pgf/number format/precision=1},
]
\addplot[draw=msdata8, fill=msfill8] coordinates {(A,49.2) (B,45.1) (C,42.3)};
\addplot[draw=msdata1, fill=msfill1] coordinates {(A,58.2) (B,55.1) (C,51.3)};
\legend{BGE-base, NIE}
\end{axis}
\end{tikzpicture}
\caption{Long-path support Recall@20 by trajectory length. Gains persist across increasingly long searches.}
\label{fig:path_length}
\end{figure}
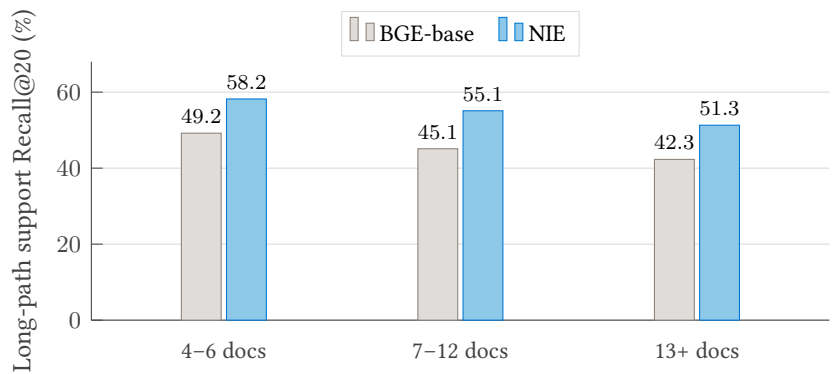

NIE provides consistent improvements across path lengths. Support Recall@20 decreases as paths lengthen for both the baseline and our model, indicating that very extended navigation sequences remain challenging.

\subsection{Robustness to Behavioral Noise}

The four-annotator audit in Section~\ref{sec:setup} finds that 26\% of sampled terminal documents are wrong under the three-of-four aggregation rule. NIE does not filter these sessions using audit judgments, yet NIE-Stop improves long-path support Recall@20 by 5.6 points over BGE and NIE-Path improves it by 8.7 points. The random-terminal control falls to 47.1, close to the unadapted score of 46.7, showing that the observed stopping document carries signal beyond path membership. Together, the audit and controls show that terminal supervision is noisy at the example level but informative in aggregate.

\subsection{Transfer to HotpotQA}
\label{sec:hotpotqa}

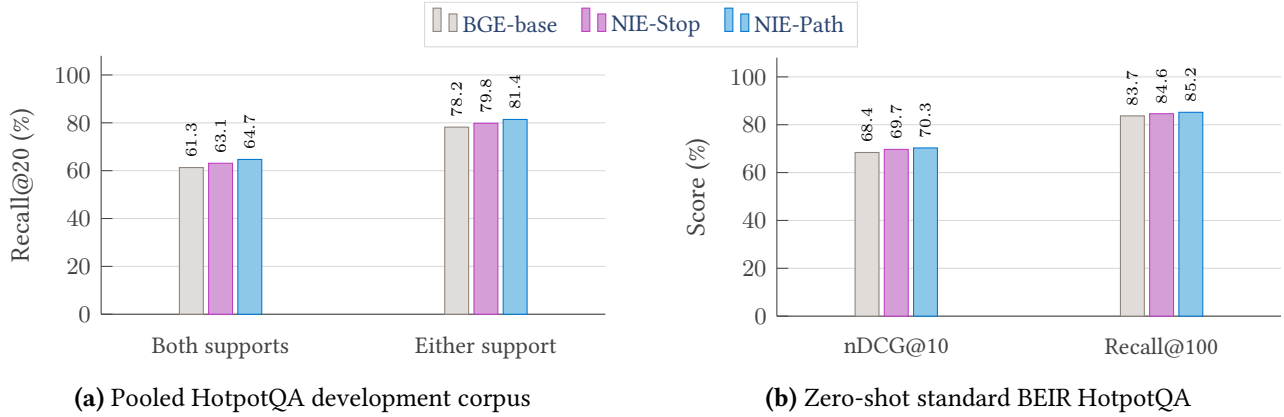
\begin{figure}[t]
\centering
\ref{transferlegend}\\[2pt]
\begin{subfigure}[b]{0.48\linewidth}
\centering
\begin{tikzpicture}
\begin{axis}[
  msgroup,
  width=\linewidth, height=5.0cm,
  ymin=0, ymax=108,
  ytick={0,20,40,60,80,100},
  ylabel={Recall@20 (\%)},
  symbolic x coords={both,either},
  xtick=data,
  xticklabels={{Both supports}, {Either support}},
  bar width=9pt,
  enlarge x limits=0.45,
  nodes near coords style={font=\tiny, rotate=90, anchor=west,
      /pgf/number format/fixed, /pgf/number format/precision=1},
  legend to name=transferlegend,
  legend style={font=\footnotesize, draw=mslightgray, fill=white,
      legend columns=-1,
      /tikz/every even column/.append style={column sep=7pt}},
]
\addplot[draw=msdata8, fill=msfill8] coordinates {(both,61.3) (either,78.2)};
\addplot[draw=msdata2, fill=msfill2] coordinates {(both,63.1) (either,79.8)};
\addplot[draw=msdata1, fill=msfill1] coordinates {(both,64.7) (either,81.4)};
\legend{BGE-base, NIE-Stop, NIE-Path}
\end{axis}
\end{tikzpicture}
\caption{Pooled HotpotQA development corpus}
\label{fig:hotpotqa}
\end{subfigure}\hfill
\begin{subfigure}[b]{0.48\linewidth}
\centering
\begin{tikzpicture}
\begin{axis}[
  msgroup,
  width=\linewidth, height=5.0cm,
  ymin=0, ymax=108,
  ytick={0,20,40,60,80,100},
  ylabel={Score (\%)},
  symbolic x coords={ndcg,r100},
  xtick=data,
  xticklabels={{nDCG@10}, {Recall@100}},
  bar width=9pt,
  enlarge x limits=0.45,
  nodes near coords style={font=\tiny, rotate=90, anchor=west,
      /pgf/number format/fixed, /pgf/number format/precision=1},
]
\addplot[draw=msdata8, fill=msfill8] coordinates {(ndcg,68.4) (r100,83.7)};
\addplot[draw=msdata2, fill=msfill2] coordinates {(ndcg,69.7) (r100,84.6)};
\addplot[draw=msdata1, fill=msfill1] coordinates {(ndcg,70.3) (r100,85.2)};
\end{axis}
\end{tikzpicture}
\caption{Zero-shot standard BEIR HotpotQA}
\label{fig:beir}
\end{subfigure}
\caption{Transfer of the frozen trajectory-adapted checkpoints. The two panels use
different corpora, protocols, and metrics; absolute values are not comparable
across panels.}
\label{fig:transfer}
\end{figure}

To evaluate generalization beyond the primary target benchmark, we test on pooled HotpotQA retrieval (Figure~\ref{fig:transfer}a). The experiment isolates transfer of the already-trained adapter by comparing the BGE initializer with its trajectory-adapted variants under identical preprocessing. Recall@20 measures retrieval of the two gold supporting passages from the global corpus constructed in Section~\ref{sec:setup}. This provides a transfer test with independently annotated multi-hop support passages.

NIE improves HotpotQA retrieval by 3.4 points for retrieving both supporting passages and 3.2 points for retrieving either passage. NIE-Stop improves over BGE-base, and NIE-Path adds a further 1.6 points on the stricter ``both'' metric. This pattern matches the target-domain results: stopping-document adaptation provides the larger gain, while the complete path objective contributes additional multi-document retrieval accuracy.

\subsection{Standard Public-Benchmark Transfer}

The pooled HotpotQA experiment above measures recovery of both or either supporting passage at rank 20 under the pooled development-corpus construction. We additionally evaluate the same frozen checkpoints on standard BEIR HotpotQA, using its released corpus, test queries, qrels, and evaluation script without task-specific fitting (Figure~\ref{fig:transfer}b). This second evaluation tests the adapter under a standard public retrieval protocol and ranking-sensitive metric; its absolute values are not directly comparable with the pooled Recall@20 results.

NIE-Stop improves nDCG@10 by 1.3 points and Recall@100 by 0.9 points over BGE-base. NIE-Path adds a further 0.6 points on each metric, for total gains of 1.9 nDCG@10 and 1.5 Recall@100 over the initializer. A paired query-bootstrap interval for the total nDCG@10 gain is $[0.8, 3.0]$ points. The public result thus mirrors the target-domain decomposition: stopping-document adaptation provides the larger increment, while path supervision adds a smaller consistent improvement to the same frozen adapter.

\subsection{Cost Analysis}

Table~\ref{tab:cost} compares the cost structure of NIE against alternative approaches for generating retrieval training data.

\begin{table}[t]
\centering
\caption{Cost comparison for retrieval training data generation. NIE avoids explicit relevance labeling when suitable agent trajectories already exist.}
\label{tab:cost}
\begin{tabular}{lccc}
\toprule
Approach & Data Source & Explicit Labels & Relative Cost \\
\midrule
Human annotation & Curated & Human & High \\
Crowdsourcing & Curated & Crowd & Medium \\
Synthetic (InPars) & Generated & LLM generation & Low \\
\midrule
\textbf{NIE (ours)} & Existing traces & \textbf{None} & \textbf{Low incremental} \\
\bottomrule
\end{tabular}
\end{table}

\paragraph{Data collection.} Our cost categorization assumes a setting in which agent navigation traces are already captured by an existing LLM-based QA system, making supervision a byproduct of normal operation rather than a newly generated dataset. In such settings, the incremental cost of supervision is dominated by storage, filtering, and training rather than by new annotation campaigns. When trajectories are generated solely for training, their LLM inference cost becomes part of the data-generation budget.

\paragraph{Labeling.} Unlike human-annotation pipelines, NIE requires no explicit relevance labels for training. Relevance signal derives entirely from agent behavior: terminal documents serve as soft positives, earlier path documents as lower-ranked path items or hard contrasts. When suitable trajectories already exist, the incremental labeling cost for NIE training is zero. Human labels remain valuable for evaluation and auditing, while the training objective is constructed from trajectories alone.

\paragraph{Training.} The reported adapter updates a base embedding model for three epochs over 2,197 trajectory-derived examples. This stage requires no new LLM inference or relevance annotation; its incremental inputs are the retained traces and embedding-model optimization.

\paragraph{Distribution-shift adaptation.} The experiment evaluates one offline encoder update across disjoint source and target datasets. Continuous adaptation is enabled by applying the same objective to successive windows of search traces, without changing the supervision interface or introducing a labeling step. The reported results establish the source-to-target update that this process repeats; they do not evaluate a multi-round online deployment. If a workflow stops retrieving whole classes of required evidence, renewed data collection, workflow changes, or full retraining are required.

\section{Analysis}
\label{sec:analysis}

\subsection{Path Ordering Analysis}

We analyze query-document geometry for the 555 source trajectories that satisfy the diagnostic filters: at least four path documents remain after corpus lookup and path construction. For each trajectory we compute the Pearson correlation between document position and query similarity; under the relevance-ordering prior, later documents should exhibit higher similarity. The statistics reported below cover every trajectory passing these filters rather than selected examples.

Across the 555 paths, 338 (60.9\%) show improved correlation after training. Mean correlation increases from $0.031$ to $0.105$, a mean shift of $0.074$, and mean terminal-document rank improves by 0.28 positions. The adjacent-pair violation rate falls from 49.5\% to 44.7\%. These population diagnostics show that NIE moves the embedding space toward the observed navigation order, while the human-labeled target evaluation determines whether that change improves evidence retrieval.

\subsection{Qualitative Analysis}

Examination of long-path queries reveals recurring patterns that explain why path order can add signal beyond terminal positives:

\paragraph{Partial evidence accumulation.} Many long-path queries involve multiple partially useful sources. In these cases, earlier documents often contain background or partial evidence, while the terminal document resolves the remaining missing fact.

\paragraph{Progressive refinement.} Agents frequently begin with broad retrievals and refine based on initial results. Navigation paths capture this refinement, with later documents addressing increasingly specific aspects of the query.

\paragraph{Context accumulation.} Some queries require background context before the answer becomes interpretable. Earlier documents in the path may be useful but insufficient, which motivates treating the ordering as soft supervision rather than labeling all nonterminal documents irrelevant.

\section{Limitations}

NIE applies to settings with retained agent search traces and appropriate data-retention policies. Deployments must ensure that trace retention, filtering, and model training comply with user privacy, enterprise policy, and applicable regulation. When trajectories are generated specifically for training, their LLM inference cost becomes part of the supervision budget; in the setting studied here, trajectories are already available from the retrieval workflow.

The path-order objective uses a monotonic relevance prior: later documents tend to be more useful than earlier ones. This prior is soft. Agents can stop on a wrong document, and early documents can contain necessary supporting evidence. The terminal-document audit and support-recall evaluation directly measure consequences of this risk in the reported setting: 26\% of audited terminal documents are wrong, yet supporting-evidence coverage improves, including on long-path queries. NIE applies to workflows whose retained traces preserve useful coverage under gradual distribution drift; systematic coverage gaps require broader retraining or workflow intervention.

Earlier path documents can be useful evidence even if they were insufficient for the agent to stop. The adjacent ordering loss addresses this by imposing local preferences rather than all-pairs irrelevance. The shuffled-order and reversed-order controls quantify whether observed chronology is more useful than arbitrary or inverted chronology under the full objective.

Implicit supervision relies on sufficient query volume to expose meaningful navigation patterns. Low-traffic or cold-start settings may benefit from complementary sources of supervision, such as lightweight synthetic data generation or selective human annotation, to bootstrap learning.

Pooled HotpotQA provides a public transfer check with human-annotated supporting facts \citep{yang2018hotpotqa}. Its question-specific duplicate identifiers make absolute scores sensitive to duplicate competition and tie handling, so it is a secondary diagnostic rather than a standard full-Wikipedia retrieval result; the primary claim is the source-to-target adaptation result on the workflow benchmark.

\section{Conclusion}

We introduced Navigation-Informed Embeddings (NIE), a family of objectives for adapting dense retrievers from agent search traces. NIE-Stop learns from the observed stopping document alone, while NIE-Path adds nonterminal comparisons and ordinal constraints for extended searches. Terminal-only adaptation improves long-path support Recall@20 from 46.7 to 52.3, and the complete path objective raises it to 55.4. Together, the objectives improve target support Recall@20 from 72.2 to 78.0 overall.

NIE makes retriever adaptation trace-driven rather than annotation-driven. The reported experiment evaluates one offline update; the same construction can be applied to subsequent trace windows as workflow distributions evolve. When suitable traces are already retained, each update has zero incremental labeling cost and requires only lightweight encoder training.

The results establish retained agent trajectories as a practical supervision channel for low-cost dense-retriever adaptation and identify ordered path structure as an additional source of signal on the hardest queries.

\bibliography{references}
\bibliographystyle{plainnat}

\end{document}